\documentclass[letterpaper, 10 pt, conference]{ieeeconf}  %

\IEEEoverridecommandlockouts                              %

\title{\LARGE \bf
Swiss DINO: Efficient and Versatile Vision Framework for\\ On-device Personal Object Search
}

\author{Kirill Paramonov$^{1}$, Jia-Xing Zhong$^{1,2,*}$, Umberto Michieli$^{1}$, Jijoong Moon$^{3}$, Mete Ozay$^{1}$%
\thanks{*Research completed during internship at Samsung R\&D Institute UK.}%
\thanks{$^{1}$ Samsung R\&D Institute UK (SRUK),
        Communications House, South St, Staines, Surrey, United Kingdom
        {\tt\small \{n.surname\}@samsung.com}}%
\thanks{$^{2}$ University of Oxford,
Wellington Square, Oxford, Oxfordshire, United Kingdom
        {\tt\small jiaxing.zhong@cs.ox.ac.uk}}%
\thanks{$^{3}$Samsung Research Korea,
        Seoul R\&D Campus, 56, Seongchon-gil, Seocho-gu, Seoul, Rep. of Korea
        {\tt\small jijoong.moon@samsung.com}}%
}

\usepackage{graphics,epsfig,graphicx,subcaption}
\usepackage{amsmath,amssymb,pifont,xcolor}
\usepackage{multicol,multirow,booktabs,arydshln}
\usepackage{hyperref,cite}

\makeatletter
\def\UrlAlphabet{%
      \do\a\do\b\do\c\do\d\do\e\do\f\do\g\do\h\do\i\do\j%
      \do\k\do\l\do\m\do\n\do\o\do\p\do\q\do\r\do\s\do\t%
      \do\u\do\v\do\w\do\x\do\y\do\z\do\A\do\B\do\C\do\D%
      \do\E\do\F\do\G\do\H\do\I\do\J\do\K\do\L\do\M\do\N%
      \do\O\do\P\do\Q\do\R\do\S\do\T\do\U\do\V\do\W\do\X%
      \do\Y\do\Z}
\def\UrlDigits{\do\1\do\2\do\3\do\4\do\5\do\6\do\7\do\8\do\9\do\0}
\g@addto@macro{\UrlBreaks}{\UrlOrds}
\g@addto@macro{\UrlBreaks}{\UrlAlphabet}
\g@addto@macro{\UrlBreaks}{\UrlDigits}
\makeatother

\usepackage{xcolor}

\begin{document}

\maketitle
\thispagestyle{empty}
\pagestyle{empty}

\begin{abstract}

In this paper, we address a recent trend in the robotic home appliances to include vision systems on personal devices, capable of personalizing the appliances on the fly.
In particular, we formulate and address an important technical task of personal object search, which involves localization and identification of personal items of interest on images captured by robotic appliances, with each item referenced only by a few annotated images.
The task is crucial for robotic home appliances and mobile systems, which need to process personal visual scenes or to operate with particular personal objects (e.g., for grasping or navigation).
In practice, personal object search presents two main technical challenges.
First, a robot vision system needs to be able to distinguish between many fine-grained classes, in the presence of occlusions and clutter.
Second, the strict resource requirements for the on-device system restrict usage of most state-of-the-art methods for few-shot learning, and often prevent on-device adaptation.
In this work we propose Swiss DINO: a simple yet effective framework for one-shot personal object search based on the recent DINOv2 transformer model, which was shown to have strong zero-shot generalization properties.
Swiss DINO handles challenging on-device personalized scene understanding requirements and does not require any adaptation training.
We show significant improvement (up to 55\%) of segmentation and recognition accuracy compared to the common lightweight solutions, and significant footprint reduction of backbone inference time (up to $100\times$) and GPU consumption (up to $10\times$) compared to the heavy transformer-based solutions\footnote{Code is available at:\\\url{https://github.com/SamsungLabs/SwissDINO}.}.
\end{abstract}

\section{INTRODUCTION}

Computer vision has a pivotal role in mobile systems and home appliances to understand the surroundings and navigate in complex environments.
Scene understanding deep neural networks have obtained outstanding results and have been successfully deployed to mass-accessible personal devices: for example industrial or domestic service robots (e.g., vacuum cleaners), assistive robots, and smartphones.

Recently, increasing attention \cite{barbato2024iros,hayes2022online, michieli2023online} has been devoted on the personalization of on-device AI vision models to tackle a variety of practical use cases. 
In this work, we focus on personal item search, whereby we want robot vision systems to localize and recognize personal user classes (or \textbf{fine-grained classes}, e.g. \textit{my dog Archie}, \textit{her dog Bruno}, \textit{my favorite cup}, \textit{your favorite flower}, etc.) on scenes.
Specifically, a user provides a small number of reference images with location annotations (either a segmentation map or a bounding box) for each personal item.
Then, given a new scene, a visual system needs to \textit{i)} determine which of the personal objects are present in the scene, and \textit{ii)} provide the location (in the form of a segmentation map or bounding box) for each of the personal objects present in the scene.
This task has found significant applications for personal assistants and service robots: for navigation (e.g., \textit{reach my white sofa}), HRI (e.g., \textit{find my dog Archie}), grasping (e.g., \textit{bring me my phone}), etc.

\begin{figure*}[tpb]
  \centering
  \includegraphics[trim=0.5cm 9cm 5.5cm 0.5cm, clip, width=1\linewidth]{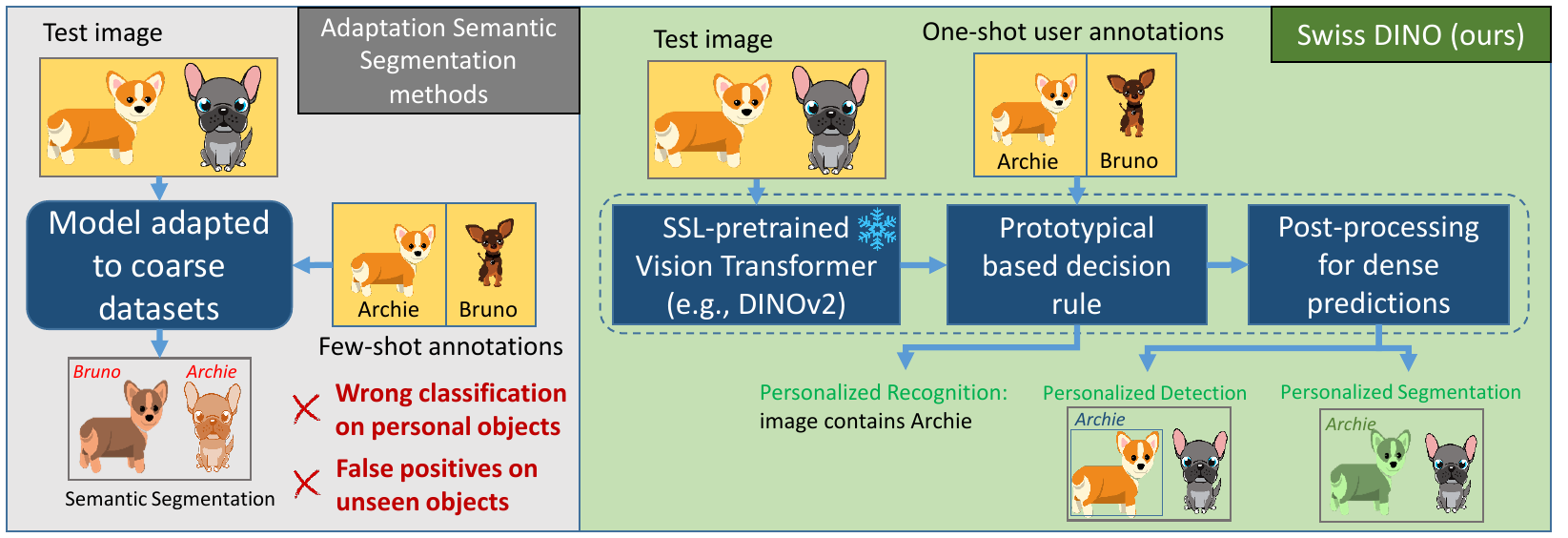}
  \caption{Comparison with semantic segmentation methods. Left: common adaptive semantic segmentation methods are adapting models to coarse datasets and do not account for multiple personal objects or unseen personal objects on a scene, thus generating false positive errors. Right: our Swiss DINO avoids false positive errors by performing open-set classification on parts of the image prior to generating segmentation masks.}
  \label{fig:problem_formulation}
\end{figure*}

Previous works have focused on different aspects of the task.
The closest comparison task to ours is the few-shot semantic segmentation \cite{catalano2023few}, which aims to segment an object on the scene given a provided reference image and mask.

In this paper, we aim to address the following limitations of existing few-shot semantic segmentation methods.
First, existing solutions only evaluate the IoU metric for the mask corresponding to the ground truth class on the image, thus not accounting for the multi-class scenario. Second, they require adaptation training on coarse datasets, thus making fine-grained classes indistinguishable in the feature space (which is part of the effect known as the \textbf{neural collapse} \cite{papyan2020collapse}).
Third, current transformer-based solutions rely on large foundation models (e.g., SAM), which may be too costly for on-device implementation.

In this work, we develop a problem statement and metrics for the personal object search task, that is closely related to the practical scenarios.
We develop a novel method for the task, which does not rely on coarse dataset training and is very lightweight, allowing seamless implementation on device.

Inspired by works showing the great versatility of the DINOv2 model \cite{oquab2023dinov2} for downstream tasks \cite{liu2024matcher, oquab2023dinov2}, we employ DINOv2 as our backbone.
Our system is called Swiss DINO, after the Swiss Army Knife, for its incredible versatility and adaptability.
Fig.~\ref{fig:problem_formulation} shows our approach and its novelty compared to existing solutions.

Our evaluation focuses on multi-instance personalization (i.e., adaptation to multiple personal objects) via one-shot transfer on multiple tasks (image classification, object detection and semantic segmentation) and datasets (iCubWorld \cite{fanello2013CVPRws} and PerSeg \cite{zhang2024personalize}).
For one-shot segmentation task, Swiss DINO improves the memory usage by up to $10\times$ and backbone inference time by up to $100\times$ compared to few-shot semantic segmentation competitors based on foundational models, while maintaining similar segmentation accuracy, as well as improving segmentation accuracy of lightweight solutions on cluttered scenes by $46\%$. %
To evaluate multi-instance identification accuracy, we adopt metrics from open-set recognition task \cite{neal2018open}.
We adapt existing segmentation methods for the multi-instance setup and show that compared to lightweight competitors pre-trained on coarse classes, Swiss DINO achieves $55\%$ identification improvement on simple scenes, and $42\%$ improvement on cluttered scenes.

The remainder of the paper is organized as follows: Sec.~\ref{sec:related} positions our paper in the current landscape of personalized scene understanding, Sec.~\ref{sec:problem} formalizes our problem setup, Sec.~\ref{sec:method} presents the details of our method, Sec.~\ref{sec:results} shows the results on several benchmark, finally Sec.~\ref{sec:conclusion} draws the conclusion of our work.

\section{Related Works}
\label{sec:related}

\noindent \paragraph{Few-Shot Semantic Segmentation} %

While early works on few-shot semantic segmentation resorted to fine-tuning large parts of models \cite{michieli2019incremental,she2020openloris,frey2022continual,michieli2021continual}, recent approaches are based on sparse feature matching \cite{xu2022aircode} or training adaptation layers with the prototypical loss \cite{cermelli2021prototype,chen2022apanet,dong2018few,li2021adaptive,wang2019panet}.
These latter approaches compute class prototypes as the average embedding of all images of a class. The label of a new (query) image is predicted by identifying the nearest prototype vector computed from the training (support) set.
The training and evaluation are usually performed on popular segmentation datasets with \textbf{coarse-level classes} (e.g. \textit{person}, \textit{cat}, \textit{car}, \textit{chair}), namely $\mathrm{PASCAL}-5^i$ \cite{shaban2017one} and $\mathrm{COCO}-20^{i}$~\cite{nguyen2019feature}.

Recent advancements in large vision models have led to novel few-shot scene understanding works, especially applied to semantic segmentation, such as PerSAM~\cite{zhang2024personalize}, and Matcher~\cite{liu2024matcher}.
PerSAM, a training-free approach, uses a single image with a reference mask to localize and segment target concepts.
Matcher, utilizing off-the-shelf vision foundation models, can showcase impressive generalization across tasks. On the other hand, both approaches are computationally expensive and not applicable on low-resource devices. %

\paragraph{Object Detection Datasets for Robotic Applications} 

Object detection and fine-grained identification are crucial tasks for robotic manipulators~\cite{mitash2023armbench}.
To boost the development of object detection methods, several datasets have been introduced.
In particular, iCubWorld~\cite{fanello2013CVPRws} is a collection of images recording the visual experience of the iCub humanoid robot observing personal user objects in its typical environment, such as a laboratory or an office.
CORe50~\cite{lomonaco2017core50} further enriches the field, offering a new benchmark for continuous object recognition, designed specifically for real-world applications such as fine-grained object detection in robot vision systems.
These datasets align with our task as they represent scenarios where few-shot personalization can be used to enhance the robot’s ability to recognize new or fine-grained objects, serving as practical representations of the use cases where our method can be applied. %
It was shown that the common classification architectures trained on coarse-level datasets have low accuracy on aforementioned datasets with fine-grained classes~\cite{pasquale2019we} when applied out-of-the-box.
This is due to the fact that the fine-grained classes become indistinguishable in the feature space after long training on coarse class classification, part of the effect known as \textbf{neural collapse}~\cite{papyan2020collapse}.
Therefore, fine-tuning~\cite{pasquale2019we} or adaptation~\cite{li2021adaptive,cermelli2021prototype} methods are often employed to separate the feature vectors for fine-grained datasets.

\paragraph{Pre-trained DINOv2 as An All-purpose Backbone}  %

In self-supervised learning (SSL), significant contributions have been made to the development of pre-trained models, such as DINO~\cite{caron2021emerging} and DINOv2~\cite{oquab2023dinov2}.
These models have demonstrated remarkable capabilities in feature extraction and object localization, making them highly transferable to our task of few-shot personalization.
Siméoni et al.\ presents a method named LOST~\cite{simeoni2021localizing} to leverage pre-trained vision transformer features for unsupervised object localization.
Melas-Kyriazi et al.~\cite{melaskyriazi2022deep} reframed image decomposition as a graph partitioning problem, using eigenvectors from self-supervised networks to segment images and localize objects.
These methods not only provide a strong foundation for our few-shot personalization method but also highlight the potential of SSL transformer backbones in overcoming the challenge of neural collapse. 

\section{PROBLEM STATEMENT}
\label{sec:problem}

In this section, we present the problem formulation of personal object search and notation for each of the three stages involved.

\subsubsection{Pre-training Stage}
\label{subsubsec:pretrain}
The first stage is to pre-train a backbone model on a large dataset.
The backbone should provide localization information of objects on an image, and have a strong ability to transfer to new personal classes, in particular avoiding neural collapse of generated features.

\subsubsection{On-device Personalization Stage}
\label{subsubsec:personalization}
After the system is implemented on a mobile or robotic device (e.g., a robot vacuum cleaner or service robot), it is shown a few images of personal objects, together with their label (e.g., \textit{dog Archie}, \textit{dog Bruno}, \textit{my mug}, etc.), and a prompt indicating the location of the object on the image, in the form of a bounding box or a segmentation map.
Those images are also known in the few-shot literature as \textit{support} images.

Although our setup can be applied to any number of support images per personal object, to simplify evaluation and notations, for the rest of the paper, we consider the most challenging one-shot setup, i.e., we get a single support image $S_c$ for each personal object index $c = 1, 2, \ldots, C$, where $C$ is the number of personal objects.

\subsubsection{On-device Open-set Personal Object Segmentation, Detection, and Recognition}
\label{subsubsec:inference}

During the on-device inference stage, we are given a new test image $Q$ (also known in the literature as the \textit{query} image).
For this image, we need to determine: \textit{i)} which personal objects, if any, are present on the image; \textit{ii)} for each of the personal objects present on the image, find its location in the form of segmentation map or a bounding box.

More formally, we define a personal object search method POS by
\begin{equation}
    \mathrm{POS}(Q) := (oloc_1(Q), \ldots, oloc_C(Q)),
\end{equation}
\begin{equation}
    oloc_c := 
    \begin{cases}
        loc_c(Q),& \text{if the object } c \text{ is present}\\
        \text{None},              & \text{otherwise}
    \end{cases}
\end{equation}
where $loc_c(Q)$ can take the form of a bounding box or a segmentation map for the object $c$ on the image $Q$.

\section{METHODOLOGY}
\label{sec:method}

\begin{figure*}[tpb]
  \centering
  \includegraphics[trim=1cm 8.3cm 1cm 0.5cm, clip, width=\textwidth]{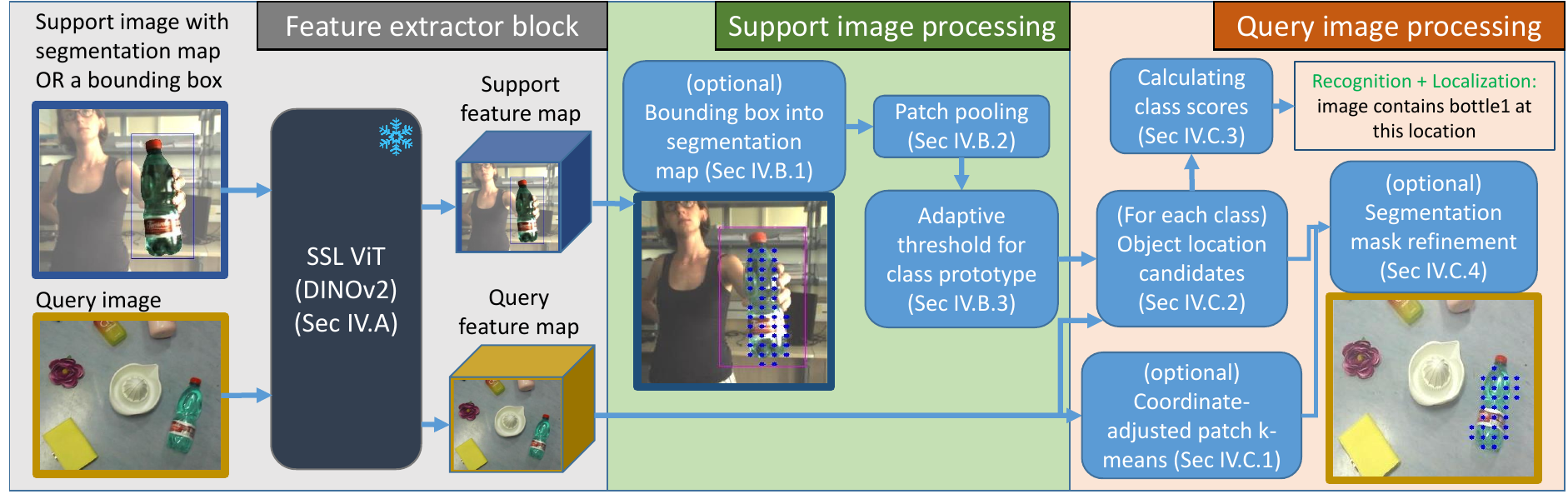}
  \caption{High-level overview of our Swiss DINO system.}
  \label{img:system_overview}
\end{figure*}

Our Swiss DINO system consists of three main components: \textit{i)} patch-level feature map extraction; \textit{ii)} support feature map processing; \textit{iii)} query feature map processing. An overview of our Swiss DINO is shown in Fig.~\ref{img:system_overview}.

\subsection{Patch-level feature map extractor}
\label{subsec:feature_extractor}

We utilize a pre-trained transformer-based patch-level feature extractor.
Inspired by the previous work~\cite{oquab2023dinov2} on the DINOv2 model, making use of its localization and fine-grained separation capabilities of its feature map, we choose DINOv2 as our transformer backbone (for a comparison between different backbone models, see Section~\ref{subsubsec:backbones}).

The backbone $\mathcal{B}$ takes an image $X$ as an input and produces \textit{i)} patch-wise feature map $X^F = (X^F_{1,1}, \ldots, X^F_{N_P,N_P})$, where $N_P$ is the number of patches along each side of the image, and $X^F_{i,j}$ is the $D$-dimensional vector corresponding to the $(i,j)$-th spatial patch on the image; and \textit{ii)} a $D$-dimensional class token $X^C$, such that $\mathcal{B}(X) = (X^F, X^C)$.

Given support images $\{S_c\}_{c=1}^C$ for each personal-level class and a query image $Q$, we compute the corresponding feature maps $\{S_c^F\}_{c=1}^C$ and $Q^F$.

\subsection{Support feature map processing}

For each personal class $c=1,2, \ldots, C$, we apply the same processing steps to the feature map $S_c^F$.
In the following, we drop the index $c$ to make the notation less cluttered.

\subsubsection{(optional) Bounding box into segmentation map}

If we are given the ground truth bounding box $b$ for the support image $S$, we consider the union of all patches $P_{i,j}$ that have non-empty intersection with $b$, denoted by $b^P$, as well as patches bordering $b^P$, denoted by $\partial b^P$.
We then partition a set of corresponding feature vectors $\{S^F_{i,j} | P_{i,j} \in b^P \cup \partial b^P \}$ into $k_S$ clusters using the $k$-means method (in our implementation, $k_S$ was empirically chosen to be $k_S=5$), denoting the set of patches in each cluster as $\mathcal{K}_r^P$, with $r=1,2,\ldots,k_S$.

Given that the patches from $\partial b^P$ are outside of the bounding box, and thus do not belong to the object of interest, we filter out the patch clusters which contain those `negative' patches, thus resulting in an (approximate) segmentation map:
\begin{equation}
    \label{eq:seg_from_bb}
    seg = \bigcup_r \{ \mathcal{K}_r^P \ | \ \partial b^P \cap \mathcal{K}_r^P = \emptyset\}.
\end{equation}

This process allows us to separate the object of interest within the bounding box from the background.

\subsubsection{Patch pooling from segmentation map}

Given a (ground truth or approximate) segmentation map $seg$ of the support image, we pick the patches $P_{i,j}$ which partially intersect with $seg$, denoting the set of those patches as $seg^P$.

We compute the patch prototype with simple average over patches in $seg^P$ by
\begin{equation}
    \label{eq:patch_proto}
    proto := avg(S^F_{i,j} | P_{i,j} \in seg^P).
\end{equation}

\subsubsection{Adaptive threshold for class prototype}

To pick the patches of interest from the query object, we choose a feature distance metric and a threshold to determine patches of interest on the query image.
As a distance metric between feature vectors, we pick the widely used cosine similarity metric.
To determine the distance threshold, we use the information about positive and negative patches on the support image.

More concretely, we denote $seg^P$ to be a set of patches that have non-empty intersection with the segmentation map $seg$, and $nseg^P$ to be a set of patches that have empty intersection with $seg$.
We compute the set of positive patch distances and the set of negative patch distances
\begin{equation}
    pd = \{ dist(S^F_{i,j}, \ proto) \ | \ P_{i,j} \in seg^P \},
\end{equation}
\begin{equation}
    nd = \{ dist(S^F_{i,j}, \ proto) \ | \ P_{i,j} \in nseg^P \}.
\end{equation}

We also remove some possible patch outliers (by removing the highest 5 percent from $pd$ and the lowest 5 percent from $nd$) to find positive and negative thresholds by ${ptr = percentile(pd, 95)}$ and $ntr = percentile(nd, 5)$.
The final adaptive threshold is taken as the minimum of positive and negative thresholds by $tr := \min(ptr, ntr)$.

\subsection{Query feature map processing}
\label{subsec:query_processing}

Given a tuple $(proto_c,\ tr_c)$ for each personal class $c=1, \ldots, C$ and query feature map $Q^F$, we use the following steps to find the patches belonging to the objects of interest.

\subsubsection{(optional for refined segmentation map) Coordinate-adjusted patch k-means}
\label{subsubsec:kmeans_refine}

First, agnostic to the set of support classes, we perform a pre-processing step on the query feature map $Q^F$.
To group together the patches corresponding to the same object, we apply $k$-means to the patch feature vectors.
In addition, we augment feature vectors with spacial information to reinforce the connectivity of patch clusters:
\begin{equation}
    Q^{F, aug}_{i,j} := concat(Q^F_{i,j}, \ \alpha_{co} i/N_P, \ \alpha_{co} j/N_P),
\end{equation}
where $\alpha_{co}$ is a coordinate scaling factor aiming to control the effect of spatial information on the resulting clusters.

We cluster the augmented patch features $Q^{F,aug}$ into $k_Q$ clusters $\mathcal{K}^Q_r$, $r=1,\ldots, k_Q$ and save those clusters for the segmentation map refinement step later.

\subsubsection{Object location candidates}
\label{subsubsec:query_loc_cands}

For each personal class $c=1, \ldots, C$, we find patches in $Q^F$ which are close enough to the class prototype, resulting in a set of patches we denote as $pseg_c^{raw}$:
\begin{equation}
    pseg^{raw}_c := \big\{ P_{i,j} \ | \ dist(Q^F_{i,j}, \ proto_c) < tr_c \big\},
\end{equation}
where $dist$ is the cosine similarity between feature vectors.

If $pseg^{raw}_c$ is empty, we choose the patch which is closest to the prototype: $pseg^{raw}_c = \arg\min_{P_{i,j}} dist(Q^F_{i,j}, \ proto_c)$.

To account for cluttered scenes with similar objects, we split $pseg^{raw}_c$ into $L$ connected subsets $(pseg_c^1, \ldots, pseg_c^L)$, thus generating $L$ candidates for the location of the object $c$ on the image.

\subsubsection{Calculating class scores}

For each candidate set of patches $pseg_c^l$, $l=1,\ldots,L$, we find the class score via patch prototype distance with support image:
\begin{equation}
    score_{c}^l := dist\big(avg(Q^F_{i,j} \ | \ P_{i,j} \in pseg_c^l), \ proto_c\big).
\end{equation}
We then choose the candidate $l_{max}$ with the maximum class score as the predicted segmentation map 
\begin{equation}
    \mathrm{pseg}_c := \bigcup_{i,j} \ \big\{P_{i,j} \ | \ P_{i,j} \in pseg_c^{l_{max}}\big\},
\end{equation}
and classification score $score_c := \max_l score_{c}^l$.

From the class score $score_c$, we determine whether object $c$ is on the image.
Similar to other score-based approaches in open-set classification \cite{neal2018open}, a classification threshold needs to be selected for a given dataset to control which predicted masks $\mathrm{pseg}_c$ we would accept, and which ones we would reject.
In the actual implementation, the classification threshold needs to be selected for each scenario empirically, while in this work we measure the capability of the method to separate positive examples from negative examples via score precision metric (see Section~\ref{subsec:metrics}).

\subsubsection{(optional) Segmentation map refinement}

While we can use the $\mathrm{pseg}_c$ as a segmentation map for the object of interest, the map usually covers only part of the object or contains holes.
To capture the whole object, we refine the patches from $\mathrm{pseg}_c$ with clusters $\mathcal{K}^Q_r$ obtained from the previous step using $k$-means by
\begin{equation}
    \mathrm{pseg}^{ref}_{c} := \bigcup_r \big\{ \mathcal{K}^Q_r \ | \ \mathcal{K}^Q_r \cap \mathrm{pseg}_c \neq \emptyset \big\}.
\end{equation}

\subsubsection{(optional) Bounding box from segmentation map}
\label{subsubsec:query_bb_from_segmap}

We can also generate a detection bounding box using the refined segmentation map $\mathrm{pseg}^{ref}_c$ by taking the extreme coordinates of the segmentation map.

\section{EXPERIMENTS}
\label{sec:results}

\subsection{Datasets}

In this section, we describe the datasets used for the evaluation of our framework.
We specifically choose datasets which \textit{i)} contain images of personal objects with different position/scale/background/lighting variations with fine-level class annotations, and \textit{ii)} include either a segmentation map or bounding box annotations.

\subsubsection{PerSEG}

The PerSEG dataset \cite{zhang2024personalize} is a convenient choice for one-shot segmentation tasks due to a collection of 40 personalized classes and high-quality segmentation maps.
The images contain salient objects that take a large part of the image and with simple, non-cluttered backgrounds, making the segmentation and classification task easier compared to other, noisier, datasets.
For few-shot evaluation, we take the first image for each class as a reference image, and test one-shot open-set classification and segmentation on the rest of the images in the class, following \cite{zhang2024personalize}.

\subsubsection{iCubWorld}

The iCubWorld dataset \cite{fanello2013CVPRws} is aimed specifically for robotics application of fine-grained object identification.
The dataset contains images from several sessions where a single object is being moved in hand across the scene, and several additional sessions where various objects are filmed in cluttered environments.

We take the subset of the sessions within dataset that contain bounding box annotations, namely \textit{i)} MIX sessions where 50 personal objects are captured in various poses, scales, and lighting conditions, one session per object, and \textit{ii)} TABLE, FLOOR1, FLOOR2, SHELF sessions where a subset of personal objects are scattered on the same scene (altogether 19 personal objects are included in those sessions).
For the evaluation of our framework, we take the first image of the MIX session as the support image.
We evaluate detection and open-set classification accuracy separately on the collection of MIX sessions (called iCW-single here) and the collection of cluttered sessions TABLE, FLOOR1, FLOOR2, SHELF (called iCW-cluttered here).

\subsection{Metrics}
\label{subsec:metrics}

In this section, we present metrics for the personal object search task, which include a localization metric and two open-set identification metrics.

The precise definitions of the metrics are to follow.

\textit{i)} To measure localization, we employ the common mIoU metric \cite{csurka2013good} between ground truth localization and predicted localization for a given personal item on the image:
\begin{equation}
    \mathrm{mIoU} := avg_i\Big(IoU\big(\mathrm{pseg}^{ref}_{gtc_i}(Q_i), \ \mathrm{gtseg}_i\big)\Big),
\end{equation}
where $(Q_i, gtc_i, \mathrm{gtseg}_i)$ are triplets of reference image, index of a personal object on the image, and ground truth localization (segmentation map for PerSEG and bounding box for iCubWorld) of the object on the image, respectively (for cluttered scenes, different personal objects on the same image would correspond to different tuples with same $Q_i$).

\textit{iii)} To measure identification accuracy (denoted by ACC), we check that the predicted score for the ground truth class is the highest among candidate locations of comparison classes near the ground truth location:
\begin{equation}
    \mathrm{ACC} := avg_i\Big(acc\big(\arg\max_c(score_c^{l_{loc}}(Q_i))\big)\Big),
\end{equation}
where $score_c^{l_{loc}}(Q_i)$ is the score of the location candidate of personal class $c$ with the highest intersection with ground truth map $\mathrm{gtseg}_i$ (the score is 0 if there is no intersecting candidate).

\textit{iii)} To measure open set identification accuracy, we employ the Average Precision metric across class scores (denoted by cPREC), which measures how well the class scores for positive examples are separated from the scores from negative examples:
\begin{equation}
    \mathrm{cPREC} := avg_c\Big(AP_i\big(score_c^{l_{loc}}(Q_i)\big)\Big).
\end{equation}

We also compare the footprints of the methods in the form of \textit{i)} inference time: how much time does it take for a backbone to process a single image; and \textit{ii)} GPU memory consumption (vRAM): how much GPU memory is required to pass a single image through the backbone, without gradients.

Since the pre- and post-processing steps are done on the CPU, the timings for those steps depend on I/O throughput and specific implementation of those steps.
However, since the k-means pre-processing step takes a considerable amount of time in Swiss DINO, we discuss the impact of k-means on the time footprint in Section \ref{subsubsec:k-means}.

\subsection{Experimental setup}

For our experiments, we use DINOv2 backbone (version without registers), with input resized to 448x448 resolution, patch size 14, and patch number $N_P=32$.

To measure the footprints, we use NVIDIA A40 single GPU, with batch size 1 during inference.

For segmentation refinement hyperparameters from Section~\ref{subsubsec:kmeans_refine}, we empirically chose $k_Q=30, \alpha_{co}=200$ for the iCubWorld dataset, and $k_Q=150, \alpha_{co}=200$ for the PerSEG dataset.
We employ efficient k-means++ method~\cite{arthur2007k} to speed up the clustering step.

\begin{table*}[t]
\setlength{\tabcolsep}{3.5pt}
    \centering
    \caption{Results on the iCubWorld dataset for the object detection task.}
    \label{tab:results:icubworld}
    \begin{tabular}{cccccccccc}
    \toprule
        \multirow{2}{*}{\textbf{Method}} & \multirow{2}{*}{\textbf{Backbone}} & \multicolumn{2}{c}{\textbf{mIoU} $\mathbf{\uparrow}$} & \multicolumn{2}{c}{\textbf{cPREC} $\mathbf{\uparrow}$} & \multicolumn{2}{c}{\textbf{ACC} $\mathbf{\uparrow}$} & \multirow{2}{*}{\textbf{Time (ms)} $\mathbf{\downarrow}$} & \multirow{2}{*}{\textbf{vRAM (MB)} $\mathbf{\downarrow}$} \\
        & & single & cluttered & single & cluttered & single & cluttered & & \\
        \midrule
        & YOLOv8-s & 54.2 & 6.5 & 8.1 & 10.6 & 9.5 & 11.6 & 7.8 & 390 \\
        YOLOv8-seg \cite{yolov8_ultralytics} & YOLOv8-m & 56.0 & 8.2 & 10.8 & 13.4 & 11.2 & 17.0 & 12.6 & 520 \\
        & YOLOv8-l & 53.1 & 7.6 & 10.8 & 10.4 & 9.6 & 6.0 & 12.2 & 676 \\
        \midrule
        & DINOv2 (ViT-s) & 65.7 & 49.8 & 61.1 & 54.8 & 46.8 & 67.3 & \textbf{7.3} & \textbf{152} \\
        Swiss DINO (ours) & DINOv2 (ViT-b) & 68.7 & 50.3 & 62.5 & \textbf{55.3} & 65.1 & \textbf{69.1} & \textbf{7.3} & 444 \\
        & DINOv2 (ViT-l) & \textbf{69.9} & \textbf{53.4} & \textbf{65.7} & 52.3 & \textbf{68.2} & 68.7 & 14.6 & 1250 \\
        \midrule
        & DINOv2 (ViT-s) & - & - & 68.9 & 96.0 & 70.8 & 93.0 & \textbf{7.3} & 152 \\
        DINOv2 bbox oracle (upper bound) & DINOv2 (ViT-b) & - & - & 68.5 & 96.7 & 72.4 & 92.9 & \textbf{7.3} & 444 \\
        & DINOv2 (ViT-l) & - & - & 70.6 & 94.4 & 74.6 & 94.2 & 14.6 & 1250 \\
        \bottomrule
    \end{tabular}
\end{table*}

\begin{table*}[t]
    \centering
    \caption{Results on the PerSEG dataset for the semantic segmentation task.}
    \label{tab:results:perseg}
    \begin{tabular}{ccccccc}
    \toprule
        \textbf{Method} & \textbf{Backbone} & \textbf{mIoU} $\mathbf{\uparrow}$ & \textbf{cPREC} $\mathbf{\uparrow}$ & \textbf{ACC} $\mathbf{\uparrow}$ & \textbf{Time (ms)} $\mathbf{\downarrow}$ & \textbf{vRAM (MB)} $\mathbf{\downarrow}$ \\
        \midrule
        \multirow{3}{*}{YOLOv8-seg} & YOLOv8-s & 85.6 & 29.9 & 29.0 & 7.8 & 390 \\
        & YOLOv8-b & 88.3 & 40.8 & 34.5 & 12.6 & 520 \\
        & YOLOv8-l & 87.4 & 33.6 & 32.3 & 12.2 & 676 \\
        \midrule
        DINOv2+M2F \cite{oquab2023dinov2,cheng2022masked} & DINOv2 (ViT-g)+M2F & 68.5 & 46.3 & 37.7 & 1415 & 17980 \\
        PerSAM \cite{zhang2024personalize} & SAM (ViT-b) & 86.1 & 89.5 & 84.3 & 758 & 1674 \\
        PerSAM \cite{zhang2024personalize} & SAM (ViT-h) & \textbf{89.3} & 91.8 & 85.6 & 1001 & 6874 \\
        Matcher \cite{liu2024matcher} & DINOv2+SAM (ViT-h) & 76.6 & \textbf{91.9} & \textbf{86.7} & 3787 & 8670 \\
        \midrule
        \multirow{3}{*}{Swiss DINO (ours)} & DINOv2 (ViT-s) & 83.5 & 91.4 & 82.0 & \textbf{7.3} & \textbf{152} \\
         & DINOv2 (ViT-b) & 83.5 & 90.4 & 81.5 & \textbf{7.3} & 444 \\
        & DINOv2 (ViT-l) & 82.4 & 89.5 & 81.3 & 14.6 & 1250 \\
        \midrule
        \multirow{3}{*}{DINOv2 bbox oracle (upper bound)} & DINOv2 (ViT-s) & - & 99.9 & 97.6 & 7.3 & 152 \\
         & DINOv2 (ViT-b) & - & 98.8 & 96.0 & 7.3 & 444 \\
        & DINOv2 (ViT-l) & - & 98.8 & 98.0 & 14.6 & 1250 \\
        \bottomrule
    \end{tabular}
\end{table*}

\subsection{Comparison methods} 

To compare our method against the existing solutions, we focus primarily on the training-free methods for semantic segmentation or detection.
To be able to adapt semantic segmentation methods for few-shot prototype-based identification task, we choose methods that provide a feature map or a prototype vector for each predicted segmentation mask or bounding box.

\subsubsection{Matcher / PerSAM / DINOv2+M2F}

Matcher \cite{liu2024matcher} and PerSAM \cite{zhang2024personalize} are state-of-the-art training-free methods for one-shot semantic segmentation.
The methods are based on prompt engineering for a large SAM \cite{kirillov2023segment} segmentation model either based on positive-negative pairs \cite{zhang2024personalize} or on DINOv2's features \cite{liu2024matcher}.
We include Matcher and PerSAM with default SAM ViT-h backbones, as well as PerSAM with smaller SAM ViT-b backbone to compare the footprint efficiency of those methods.
We also consider DINOv2+Mask2Former, using an M2F \cite{cheng2022masked} segmentation head on top of DINOv2 \cite{oquab2023dinov2} specifically trained for segmentation on the ADE20k dataset \cite{zhou2017scene} with coarse-level classes.

To adapt and evaluate PerSAM and Matcher for multi-class identification, we \textit{i)} extract the feature map from the respective backbones (DINOv2 for Matcher, SAM for PerSAM); \textit{ii)} average the features over the support and predicted query masks to get the prototypes; and \textit{iii)} apply cosine distance to calculate $score_c(Q)$ for each query image.

To adapt DINOv2+M2F for personal object search task, we use pre-softmax feature vectors of M2F head.
We then perform the same steps as our method, but using DINOv2+M2F as an alternative backbone.

The methods above require a precise segmentation map extracted from the reference image, without good extension to bounding box annotations.
Therefore, we don't consider those methods for the iCubWorld dataset, which only provides bounding box annotation for the reference images of personal objects.

\subsubsection{YOLOv8-seg}

YOLOv8-seg is an instance segmentation model based on the state-of-the-art YOLOv8 \cite{yolov8_ultralytics} lightweight detection method, and pre-trained on COCO \cite{lin2014microsoft} dataset.
It is particularly convenient for us, since the model outputs feature vectors for each of the candidate bounding boxes and masks.

To adapt the model for the personal object search task, we \textit{i)} extract the support prototype vector for the bounding box with the highest IoU and the ground truth bounding box; \textit{ii)} find the query bounding box with the closest prototype on the query image to get IoU score; and \textit{iii)} extract the query prototype from the ground-truth query bounding box to calculate cPREC and ACC.
We apply YOLOv8-seg on the iCubWorld dataset for detection and on the PerSEG dataset for segmentation.

\subsubsection{DINOv2 bounding box oracle}

To have an upper bound reference for the identification metrics cPREC and ACC, we assume that the ground truth location of the personal object on the test image is known.
We call this method ``DINOv2 bbox oracle" which has similar computational resources.

Knowing ground truth bounding boxes, we crop support images $S_c$ and a query image $Q$ into $S_c^{bb}$ and $Q^{bb}$ respectively.
We then use DINOv2 class tokens $(S_c^{bb})^C$ and $(Q^{bb})^C$ as prototypes and compute the class score $score_c(Q)$ as the cosine distance between corresponding prototypes.
Then, we can calculate cPREC and ACC metrics using $score_c(Q)$ as utilized before.

\subsection{Main results and discussion}

\subsubsection{Results on iCubWorld}

As we see from Table~\ref{tab:results:icubworld}, Swiss DINO significantly outperforms the lightweight comparison method YOLOv8-seg on personal object detection on the iCubWorld dataset.
Swiss DINO achieves 16\%/46\% IOU improvement on single-object and cluttered scenes, respectively.
Swiss DINO also shows significant improvement in personal object identification, showing 55\%/40\% cPREC open-set score improvement, and 57\%/52\% classification accuracy improvement on single and cluttered scenes respectively.

Significant mIOU gap in cluttered scenes compared to YOLOv8-seg method is caused by the wrong bounding box picked as the prediction, and the large gap in cPREC and ACC metrics is caused by a large number of false positive predictions near the ground truth location of the object.
Given that our adaptation of YOLOv8-seg chooses the bounding box with the feature vector closest to the ground truth class prototype, this shows the poor separation of the feature vectors for fine-grained classes.

Compared to the bounding box-oracle method, on the iCubWorld-single dataset the cPREC and ACC scores of our approach are only about 5\% below the upper bound, while on the iCubWorld-cluttered dataset, we observe a more significant 25\% accuracy gap, likely due to smaller object scale and presence of similar objects on the cluttered images.

\subsubsection{Results on PerSEG}

From Table~\ref{tab:results:perseg} we see that Swiss DINO also outperforms YOLOv8-seg on semantic segmentation on the PerSEG dataset in terms of personal object identification metrics (50\% cPREC improvement) and classification accuracy (48\% improvement), while maintaining similar computational footprint and slightly smaller IoU on segmentation maps.

Compared to DINOv2+M2F, Swiss DINO shows 25\% IoU, 45\% cPREC, and 49\% ACC improvement, while using a much smaller backbone.
This again shows how fine-tuning of segmentation head on a coarse-level dataset harms discriminative properties of the features in personalized scenarios.
Compared to heavy SAM-based Matcher and PerSAM-b/h, Swiss DINO achieves $100\times$ backbone inference time speedup and $10\times$ improved GPU memory usage while maintaining competitive segmentation and identification scores.

Overall, the results show outstanding capabilities for zero-shot transfer of DINOv2 feature maps onto new tasks (i.e., segmentation and detection) and personalized classes compared to other backbones trained on large datasets, namely: CNN-based YOLOv8 architecture, specialized Mask2Former segmentation head and SAM foundation model.

\subsubsection{Impact of k-means}
\label{subsubsec:k-means}

In the on-server implementation of our method, most of the inference time is spent on the k-means pre-processing step for the query images (Step~\ref{subsubsec:kmeans_refine}), which is done on the CPU.
This is because k-means is performed on $N_P^2 = 1024$ feature vectors of $D=384/768/1024$ dimensions from vit-s/b/l backbones respectively.

The k-means method takes 0.19/0.23/0.25 seconds per query image for vit-s/b/l backbones respectively, which is about 95\% of the overall inference time, compared to about 10 milliseconds spent on query feature map extraction (Step~\ref{subsec:feature_extractor}), and 1 millisecond spent on the rest of the query post-processing (Steps~\ref{subsubsec:query_loc_cands}-\ref{subsubsec:query_bb_from_segmap}).
Our experiments were done on 64 Intel(R) Xeon(R) Gold 5218 CPU @ 2.30GHz cores.

Note that the k-means refinement step for query images is optional and does not affect identification metrics cPREC and ACC, since the class score $score_c(Q)$ is computed using the non-refined segmentation map.
Therefore, the non-refined version is perfect for applications that only need partial localization information (e.g., a single point on the object of interest).
For mIOU scores for segmentation maps without refinement, see Table~\ref{tab:ablation:hyperparams}.

Also note that even with costly refinement step, the overall inference time is still $2\times$ lower compared to heavy-weight methods like PerSAM and Matcher, while still maintaining significant gains by $10\times$ on vRAM footprint.

\begin{table}[t]
\setlength{\tabcolsep}{2.9pt}
    \centering
    \caption{Ablation on the choice of the vision transformer backbone on the iCW-single dataset. The number in parenthesis is the number of patches along the image side.}
    \label{tab:ablation:backbones}
    \begin{tabular}{ccccc}
    \toprule
        \textbf{Backbone} & \textbf{mIoU} $\mathbf{\uparrow}$ & \textbf{cPREC} $\mathbf{\uparrow}$ & \textbf{Time (ms)} $\mathbf{\downarrow}$ & \textbf{vRAM (MB)} $\mathbf{\downarrow}$ \\
        \midrule
        OpenCLIP-ViT-b (14)& 29.7 & 52.3 & 8.1 & 374 \\
        OpenCLIP-ViT-l (16) & 35.4 & 61.9 & 16.6 & 934 \\
        \midrule
        DeiT-s (14) & 39.6 & 50.3 & 6.4 & 112  \\
        DeiT-b (14) & 37.5 & 60.3 & 6.7 & 388  \\
        \midrule
        DINO-ViT-s (32) & 58.6 & 54.7 & 8.5 & 182 \\
        DINO-ViT-b (32) & 56.3 & 54.0 & 8.6 & 554 \\
        \midrule
        DINOv2-ViT-s (32) & 65.7 & 61.1 & 7.3 & 152 \\
        DINOv2-ViT-b (32) & 68.7 & 62.5 & 7.3 & 444 \\
        DINOv2-ViT-l (32) & 69.9 & 65.7 & 14.6 & 1250 \\
        \bottomrule
    \end{tabular}
\end{table}

\begin{table}[t]
\setlength{\tabcolsep}{5pt}
    \centering
    \caption{Ablation on the number of  clusters $k_Q$ and coordinate scaling $\alpha_{co}$. All experiments are done with the DINOv2 (ViT-s) backbone.}
    \label{tab:ablation:hyperparams}
    \begin{tabular}{ccccc}
    \toprule
        $k_Q$ & $\alpha_{co}$ & iCW-single mIoU & iCW-clut mIoU & PerSEG mIoU \\
        \midrule
        $\infty$ & 0 & 37.2 & 7.3 & 81.1 \\
        \midrule
        150 & 200 & 51.2 & 32.0 & \textbf{83.5} \\
        60 & 200 & 60.4 & 47.4 & 80.3 \\
        30 & 200 & \textbf{65.7} & \textbf{49.8} & 75.7 \\
        \midrule
        30 & 0 & 68.7 & 43.2 & \textbf{79.5} \\
        30 & 50 & \textbf{69.2} & 45.3 & 78.5 \\
        30 & 200 & 65.7 & \textbf{49.8} & 75.7 \\
        \bottomrule
    \end{tabular}
\end{table}

\subsection{Ablations}

\subsubsection{ViT backbone ablations}
\label{subsubsec:backbones}

In this section, we motivate the choice of DINOv2 backbone by comparing the accuracy of the method using popular self-supervised vision transformer backbones: DINO \cite{caron2021emerging}, DeiT \cite{touvron2021training} and OpenCLIP \cite{cherti2023reproducible}.

As we see from Table~\ref{tab:ablation:backbones}, DINOv2 outperforms other backbones by 7-11\% in IoU and 1-5\% in cPREC metrics, while maintaining similar or better computational footprints.

\subsubsection{Hyperparameter ablations}

In this section, we analyze the effect of hyperparameters.
In particular, we see how $k_Q$ and $\alpha_{co}$ used in segmentation mask refinement affect IoU score for the iCW-single, iCW-cluttered, and PerSEG datasets. 
From Table~\ref{tab:ablation:hyperparams}, we see that reducing the number of clusters $K_Q$ improves the IoU score by 32\%/42\% on the iCW single and cluttered respectively, while the number of clusters needs to be high on the cleaner PerSEG dataset to exclude the false positive patches.
Coordinate scaling $\alpha_{co}$ does not affect the accuracy metrics much (yielding 5\% improvement on cluttered scenes), however, we see from qualitative observations that the inclusion of coordinate scaling makes the results more robust to scene variation.
Following this result, we choose $k_Q=30, \ \alpha_{co}=200$ for the iCubWorld dataset, and $k_Q=150, \ \alpha_{co}=200$ for the PerSEG dataset.

\section{CONCLUSION}
\label{sec:conclusion}

In this work, we have introduced a novel problem and metrics for the personal object search task, which is directly related to practical robot vision tasks performed by mobile and robotic systems such as home appliances and robotic manipulators, in which the system needs to localize all present objects of interest in a cluttered scene, where each object is only referenced by a few images.

To address this task, we have introduced Swiss DINO  that leverages SSL-pretrained DINOv2's 
feature maps having strong discriminative and localization properties.
Swiss DINO presents novel clustering-based segmentation/detection mechanisms to alleviate the need for additional specialized modules for such dense prediction tasks.

We compare our framework to common lightweight solutions, as well as heavy transformer-based solutions.
We show significant improvement (up to 55\%) of segmentation and recognition accuracy compared to the former methods, and significant footprint reduction of backbone inference time (up to $100\times$) and GPU consumption (up to $10\times$) compared to the latter methods, allowing seamless implementation on robotic devices.

Altogether, this work shows the power and versatility of self-supervised transformer models on personal object search and various downstream tasks.
In future work, we plan to extend Swiss DINO for continually learning new generic as well as new personal objects.

\bibliographystyle{IEEEbib}
\bibliography{refs.bib}

\begin{thebibliography}{10}

\bibitem{barbato2024iros}
Francesco Barbato, Umberto Michieli, Jijoong Moon, Pietro Zanuttigh, and Mete
  Ozay,
\newblock ``Cross-architecture auxiliary feature space translation for
  efficient few-shot personalized object detection,''
\newblock in {\em IROS}, 2024.

\bibitem{hayes2022online}
Tyler~L Hayes and Christopher Kanan,
\newblock ``{Online Continual Learning for Embedded Devices},''
\newblock {\em CoLLAs}, 2022.

\bibitem{michieli2023online}
Umberto Michieli and Mete Ozay,
\newblock ``Online continual learning for robust indoor object recognition,''
\newblock in {\em IROS}. IEEE, 2023.

\bibitem{catalano2023few}
N.~Catalano and M.~Matteucci,
\newblock ``Few shot semantic segmentation: a review of methodologies and open
  challenges,''
\newblock {\em arXiv:2304.05832}, 2023.

\bibitem{papyan2020collapse}
Vardan Papyan, X.~Y. Han, and David~L. Donoho,
\newblock ``Prevalence of neural collapse during the terminal phase of deep
  learning training,''
\newblock {\em PNAS}, vol. 117, no. 40, pp. 24652--24663, 2020.

\bibitem{oquab2023dinov2}
Maxime Oquab, Timoth{\'e}e Darcet, Th{\'e}o Moutakanni, Huy Vo, Marc
  Szafraniec, Vasil Khalidov, Pierre Fernandez, Daniel Haziza, Francisco Massa,
  Alaaeldin El-Nouby, et~al.,
\newblock ``Dinov2: Learning robust visual features without supervision,''
\newblock {\em arXiv:2304.07193}, 2023.

\bibitem{liu2024matcher}
Yang Liu, Muzhi Zhu, Hengtao Li, Hao Chen, Xinlong Wang, and Chunhua Shen,
\newblock ``Matcher: Segment anything with one shot using all-purpose feature
  matching,''
\newblock in {\em ICLR}, 2024.

\bibitem{fanello2013CVPRws}
Sean~Ryan Fanello, Carlo Ciliberto, Matteo Santoro, Lorenzo Natale, Giorgio
  Metta, Lorenzo Rosasco, and Francesca Odone,
\newblock ``{iCub World: Friendly Robots Help Building Good Vision
  Data-Sets},''
\newblock in {\em CVPRW}, 2013.

\bibitem{zhang2024personalize}
Renrui Zhang, Zhengkai Jiang, Ziyu Guo, Shilin Yan, Junting Pan, Hao Dong,
  Yu~Qiao, Peng Gao, and Hongsheng Li,
\newblock ``Personalize segment anything model with one shot,''
\newblock in {\em ICLR}, 2024.

\bibitem{neal2018open}
Lawrence Neal, Matthew Olson, Xiaoli Fern, Weng-Keen Wong, and Fuxin Li,
\newblock ``Open set learning with counterfactual images,''
\newblock in {\em ECCV}, 2018, pp. 613--628.

\bibitem{michieli2019incremental}
Umberto Michieli and Pietro Zanuttigh,
\newblock ``Incremental learning techniques for semantic segmentation,''
\newblock in {\em CVPRW}, 2019.

\bibitem{she2020openloris}
Qi~She, Fan Feng, Xinyue Hao, Qihan Yang, Chuanlin Lan, Vincenzo Lomonaco,
  Xuesong Shi, Zhengwei Wang, Yao Guo, Yimin Zhang, et~al.,
\newblock ``Openloris-object: A robotic vision dataset and benchmark for
  lifelong deep learning,''
\newblock in {\em ICRA}. IEEE, 2020, pp. 4767--4773.

\bibitem{frey2022continual}
Jonas Frey, Hermann Blum, Francesco Milano, Roland Siegwart, and Cesar Cadena,
\newblock ``Continual adaptation of semantic segmentation using complementary
  2d-3d data representations,''
\newblock {\em IEEE RA-L}, vol. 7, no. 4, pp. 11665--11672, 2022.

\bibitem{michieli2021continual}
Umberto Michieli and Pietro Zanuttigh,
\newblock ``Continual semantic segmentation via repulsion-attraction of sparse
  and disentangled latent representations,''
\newblock in {\em CVPR}, 2021, pp. 1114--1124.

\bibitem{xu2022aircode}
Kuan Xu, Chen Wang, Chao Chen, Wei Wu, and Sebastian Scherer,
\newblock ``Aircode: A robust object encoding method,''
\newblock {\em IEEE Robotics and Automation Letters}, vol. 7, no. 2, pp.
  1816--1823, 2022.

\bibitem{cermelli2021prototype}
Fabio Cermelli, Massimiliano Mancini, Yongqin Xian, Zeynep Akata, and Barbara
  Caputo,
\newblock ``Prototype-based incremental few-shot segmentation,''
\newblock in {\em BMVC}, 2021.

\bibitem{chen2022apanet}
Jiacheng Chen, Bin-Bin Gao, Zongqing Lu, Jing-Hao Xue, Chengjie Wang, and
  Qingmin Liao,
\newblock ``Apanet: Adaptive prototypes alignment network for few-shot semantic
  segmentation,''
\newblock {\em IEEE T-MM}, 2022.

\bibitem{dong2018few}
Nanqing Dong and Eric~P Xing,
\newblock ``Few-shot semantic segmentation with prototype learning.,''
\newblock in {\em BMVC}, 2018, vol.~3.

\bibitem{li2021adaptive}
Gen Li, Varun Jampani, Laura Sevilla-Lara, Deqing Sun, Jonghyun Kim, and
  Joongkyu Kim,
\newblock ``Adaptive prototype learning and allocation for few-shot
  segmentation,''
\newblock in {\em CVPR}, 2021, pp. 8334--8343.

\bibitem{wang2019panet}
Kaixin Wang, Jun~Hao Liew, Yingtian Zou, Daquan Zhou, and Jiashi Feng,
\newblock ``Panet: Few-shot image semantic segmentation with prototype
  alignment,''
\newblock in {\em ICCV}, 2019, pp. 9197--9206.

\bibitem{shaban2017one}
Amirreza Shaban, Shray Bansal, Zhen Liu, Irfan Essa, and Byron Boots,
\newblock ``One-shot learning for semantic segmentation,''
\newblock 2017.

\bibitem{nguyen2019feature}
Khoi Nguyen and Sinisa Todorovic,
\newblock ``Feature weighting and boosting for few-shot segmentation,''
\newblock in {\em ICCV}, 2019.

\bibitem{mitash2023armbench}
Chaitanya Mitash, Fan Wang, Shiyang Lu, Vikedo Terhuja, Tyler Garaas, Felipe
  Polido, and Manikantan Nambi,
\newblock ``Armbench: An object-centric benchmark dataset for robotic
  manipulation,''
\newblock in {\em ICRA}. IEEE, 2023, pp. 9132--9139.

\bibitem{lomonaco2017core50}
Vincenzo Lomonaco and Davide Maltoni,
\newblock ``Core50: a new dataset and benchmark for continuous object
  recognition,''
\newblock in {\em CoRL}, 2017.

\bibitem{pasquale2019we}
Giulia Pasquale, Carlo Ciliberto, Francesca Odone, Lorenzo Rosasco, and Lorenzo
  Natale,
\newblock ``Are we done with object recognition? the icub robot’s
  perspective,''
\newblock {\em RAS}, vol. 112, pp. 260--281, 2019.

\bibitem{caron2021emerging}
Mathilde Caron, Hugo Touvron, Ishan Misra, Herv{\'e} J{\'e}gou, Julien Mairal,
  Piotr Bojanowski, and Armand Joulin,
\newblock ``Emerging properties in self-supervised vision transformers,''
\newblock in {\em ICCV}, 2021, pp. 9650--9660.

\bibitem{simeoni2021localizing}
Oriane Sim{\'e}oni, Gilles Puy, Huy~V Vo, Simon Roburin, Spyros Gidaris, Andrei
  Bursuc, Patrick P{\'e}rez, Renaud Marlet, and Jean Ponce,
\newblock ``Localizing objects with self-supervised transformers and no
  labels,''
\newblock {\em arXiv:2109.14279}, 2021.

\bibitem{melaskyriazi2022deep}
Luke Melas-Kyriazi, Christian Rupprecht, Iro Laina, and Andrea Vedaldi,
\newblock ``Deep spectral methods: A surprisingly strong baseline for
  unsupervised semantic segmentation and localization,''
\newblock in {\em CVPR}, 2022.

\bibitem{csurka2013good}
Gabriela Csurka, Diane Larlus, Florent Perronnin, and France Meylan,
\newblock ``What is a good evaluation measure for semantic segmentation?.,''
\newblock in {\em BMVC}, 2013, vol.~27, pp. 10--5244.

\bibitem{arthur2007k}
David Arthur, Sergei Vassilvitskii, et~al.,
\newblock ``k-means++: The advantages of careful seeding,''
\newblock in {\em Soda}, 2007, vol.~7, pp. 1027--1035.

\bibitem{yolov8_ultralytics}
Glenn Jocher, Ayush Chaurasia, and Jing Qiu,
\newblock ``{Ultralytics YOLOv8},'' 2023.

\bibitem{cheng2022masked}
Bowen Cheng, Ishan Misra, Alexander~G Schwing, Alexander Kirillov, and Rohit
  Girdhar,
\newblock ``Masked-attention mask transformer for universal image
  segmentation,''
\newblock in {\em CVPR}, 2022, pp. 1290--1299.

\bibitem{kirillov2023segment}
Alexander Kirillov, Eric Mintun, Nikhila Ravi, Hanzi Mao, Chloe Rolland, Laura
  Gustafson, Tete Xiao, Spencer Whitehead, Alexander~C Berg, Wan-Yen Lo,
  et~al.,
\newblock ``Segment anything,''
\newblock {\em ICCV}, 2023.

\bibitem{zhou2017scene}
Bolei Zhou, Hang Zhao, Xavier Puig, Sanja Fidler, Adela Barriuso, and Antonio
  Torralba,
\newblock ``Scene parsing through ade20k dataset,''
\newblock in {\em CVPR}, 2017, pp. 633--641.

\bibitem{lin2014microsoft}
Tsung-Yi Lin, Michael Maire, Serge Belongie, James Hays, Pietro Perona, Deva
  Ramanan, Piotr Doll{\'a}r, and C~Lawrence Zitnick,
\newblock ``Microsoft coco: Common objects in context,''
\newblock in {\em ECCV}, 2014, pp. 740--755.

\bibitem{touvron2021training}
Hugo Touvron, Matthieu Cord, Matthijs Douze, Francisco Massa, Alexandre
  Sablayrolles, and Herv{\'e} J{\'e}gou,
\newblock ``Training data-efficient image transformers \& distillation through
  attention,''
\newblock in {\em International conference on machine learning}. PMLR, 2021,
  pp. 10347--10357.

\bibitem{cherti2023reproducible}
Mehdi Cherti, Romain Beaumont, Ross Wightman, Mitchell Wortsman, Gabriel
  Ilharco, Cade Gordon, Christoph Schuhmann, Ludwig Schmidt, and Jenia Jitsev,
\newblock ``Reproducible scaling laws for contrastive language-image
  learning,''
\newblock in {\em CVPR}, 2023, pp. 2818--2829.

\end{thebibliography}

\end{document}